%% file: icml2024.tex
\documentclass{article}

\usepackage{microtype}
\usepackage{graphicx}
\usepackage{subfigure}
\usepackage{booktabs} 

\usepackage{hyperref}

\usepackage[accepted]{icml2024}

\usepackage{amsmath}
\usepackage{amssymb}
\usepackage{mathtools}
\usepackage{amsthm}

\usepackage[capitalize,noabbrev]{cleveref}

\usepackage{hyperref}
\usepackage{url}
\usepackage{graphicx}
\usepackage{wrapfig}
\usepackage{adjustbox}
\usepackage{booktabs}
\usepackage{colortbl}
\usepackage{arydshln}
\usepackage{comment}
\usepackage{url}
\usepackage{tcolorbox}
\usepackage{tikz}
\usepackage{graphicx,wrapfig,lipsum}
\usepackage{amsmath}
\usepackage{placeins}
\usepackage{amsmath}
\DeclareMathOperator*{\argmax}{arg\,max}

\usepackage[textwidth=2.5cm]{todonotes}

\newcommand*\circled[1]{\tikz[baseline=(char.base)]{
            \node[shape=circle,draw,inner sep=0.2pt] (char) {#1};}}

\icmltitlerunning{Submission and Formatting Instructions for ICML 2024}

\begin{document}

\twocolumn[
\icmltitle{Do Compressed LLMs Forget Knowledge? \\ An Experimental Study with Practical Implications}



\icmlsetsymbol{equal}{*}

\begin{icmlauthorlist}
\icmlauthor{Duc  Hoang}{yyy}
\icmlauthor{Minsik Cho}{comp}
\icmlauthor{Thomas Merth}{comp}
\icmlauthor{Mohammad Rastegari}{comp}
\icmlauthor{Zhangyang Wang}{yyy}
\end{icmlauthorlist}

\icmlaffiliation{yyy}{University of Texas at Austin}
\icmlaffiliation{comp}{Apple}

\icmlcorrespondingauthor{Duc Hoang}{hoangduc@utexas.edu}
\icmlcorrespondingauthor{Minsik Cho}{minsik\_cho@apple.com}
\icmlcorrespondingauthor{Thomas Merth}{tmerth@apple.com}
\icmlcorrespondingauthor{Mohammad Rastegari}{mrastegari@apple.com}
\icmlcorrespondingauthor{Zhangyang Wang}{atlaswang@utexas.edu}

\icmlkeywords{Machine Learning, ICML}

\vskip 0.3in
]



\printAffiliationsAndNotice{}  

\begin{abstract}

Compressing Large Language Models (LLMs) often leads to reduced performance, especially for knowledge-intensive tasks. In this work, we dive into how compression damages LLMs' inherent knowledge and the possible remedies. We start by proposing two conjectures on the nature of the damage: one is certain knowledge being \textbf{forgotten (or erased)} after LLM compression, hence necessitating the compressed model to (re)learn from data with additional parameters; the other presumes that knowledge is \textbf{internally displaced} and hence one requires merely ``inference re-direction" with input-side augmentation such as prompting, to recover the knowledge-related performance. Extensive experiments are then designed to (in)validate the two conjectures. We observe the promise of prompting in comparison to model tuning; we further unlock prompting's potential by introducing a variant called Inference-time Dynamic Prompting (IDP), that can effectively increase prompt diversity without incurring any inference overhead. Our experiments consistently suggest that compared to the classical re-training alternatives such as LoRA, prompting with IDP leads to better or comparable post-compression performance recovery, while saving the extra parameter size by $\mathbf{21\times}$ and reducing inference latency by \textbf{60\%}. Our experiments hence strongly endorse the conjecture of ``knowledge displaced" over ``knowledge forgotten", and shed light on a new efficient mechanism to restore compressed LLM performance. We additionally visualize and analyze the different attention and activation patterns between prompted and re-trained models, demonstrating they achieve performance recovery in two different regimes.


\end{abstract}

\section{Introduction}

Large language models (LLMs) like GPT-4 and ChatGPT have emerged as powerful tools in language generation and reasoning, pushing the boundaries of AI to rival human-like capabilities \cite{openai2023gpt4}. These advancements, however, are accompanied by significant challenges, primarily their massive size and the consequent high computational costs \cite{Chen2023FrugalGPTHT}. This has led to a growing emphasis on model compression as a means to make LLMs more accessible and efficient for broader industrial applications.

Model compression techniques, such as quantization and sparsification, have since become increasingly popular for reducing the size of LLMs without significantly compromising their performance. Traditional approaches often involve post-compression re-training to mitigate performance losses \citep{han2015deep}. More recent `training-free' compression methods, like GPTQ \citep{Frantar2022GPTQAP} and SparseGPT \citep{Frantar2023SparseGPTML}, promise minimal impact on perplexity and standard task benchmarks. Nevertheless, studies like \citet{jaiswal2023compressing} reveal that these compressed models still suffer from reduced effectiveness in \textit{knowledge-intensive} language generation or reasoning tasks.


What transpires within a compressed model that leads to diminished performance on tasks demanding extensive knowledge? Is this knowledge permanently lost in the compression process, or is it merely obscured? Addressing these questions is not solely of theoretical interest; it has tangible implications for devising strategies to effectively counteract the impacts of compression on model knowledge. To this end, we introduce two hypotheses regarding the root cause of this performance degradation: the first posits that key knowledge is \textbf{forgotten (or erased)} as a consequence of LLM compression, necessitating a re-learning process with the addition of extra parameters \cite{Hu2021LoRALA}; the second hypothesis suggests that the knowledge is merely \textbf{internally displaced} within the LLM. This implies that strategic redirection of knowledge flow, potentially through input-side enhancements like prompting \cite{Xu2023CompressTP}, could efficiently recover model accuracy. A more comprehensive exploration of these ideas is presented in Section 2.2.

We embarked on a series of extensive experiments to test two central hypotheses of compressed LLM performance loss: ``knowledge displaced” versus “knowledge forgotten”. To effectively validate these hypotheses, we scrutinized existing prompt-tuning methods and recognized their limitations, particularly the reliance on a single prompt across varied input formats and knowledge domains. This led us to propose the \textbf{Inference-time Dynamic Prompting} (IDP) approach to auto-select appropriate prompts per input. IDP sets itself apart from previous ensemble techniques \citep{2020t5, peng2023model}, offering a one-shot selection process with nearly no overhead compared to one fixed prompt, thanks to Key-Value caching. 

Our empirical findings strongly support the hypothesis of \textbf{``knowledge displaced"} over \textbf{``knowledge forgotten"} by demonstrating prompting by IDP leads to more favorable cost-effectiveness in performance recovery. As illustrated in Figure \ref{fig:param_perf}, compared to classical model re-training methods using LoRA, IDP achieved comparable or superior performance in adapting compressed LLM models across various tasks, while attaining a remarkable reduction in parameter overhead – up to \textbf{21 times}, besides reducing inference latency by \textbf{60\%}. The performance of IDP is robust even at fairly short prompt lengths (Figure \ref{fig:robust}). Our investigation into layer-wise cosine similarity (Figure \ref{fig:cos}) further revealed that, compared to baseline attention patterns, prompt-tuning via IDP leads to significant divergences, whereas re-trained models tend to align more closely with the baseline, despite achieving similar outcomes.

In summary, our contributions are:
\begin{itemize}
    \item We critically examine the impact of compression on LLMs' knowledge, formally raising the conjectures of knowledge 'displacement' versus 'forgetfulness'.
    
    \item We thoughtfully design experiments to endorse the hypothesis of \textbf{``knowledge displaced"} over \textbf{``knowledge forgotten"}. We also reveal a number of insights, including two different regimes of performance recovery.
    
    \item As the practical implication, all experimental results collectively underscore the efficacy of our \textbf{newly designed IDP method} -  achieving similar performance recovery to LoRA, at orders-of-magnitude lower parameter and latency overheads,  
\end{itemize}






\section{Background and Conjecture}



\input{figures/compression_baseline}


\subsection{LLM compression background \& caveats}
In compression, we address the challenges of size and latency inherent to LLMs by targeting the model's parameters. Broadly, compressive techniques are grouped into two main categories: compression-aware training and post-training compression. Post-training compression holds particular appeal for exceptionally large models where the costs associated with full model training or even fine-tuning can be prohibitive. Given its relevance, we narrow our discussion to this category. 

Firstly, quantization refers to reducing the model's footprint by decreasing the bit precision of its weights \citep{Frantar2022GPTQAP, yao2022zeroquant, Xiao2022SmoothQuantAA}. Quantization not only shrinks the model's size but also accelerates inference, as operations over lower-precision weights are computationally less demanding. Secondly, sparsification, often referred to as pruning, revolves around the concept of selectively removing certain weight elements or masking activation values \citep{Frantar2023SparseGPTML, Hubara2021AcceleratedSN, Hubara2021AccuratePT}. The objective is to trim the less salient portions of the model, thereby reducing computational overhead or enhancing model throughput. 

Using GPTQ and SparseGPT for model compression, Figure \ref{fig:compression_baseline} shows a performance drop when
lowering bit counts or parameters, except for the int8 quantization. This trend aligns with the claims by \citet{Frantar2022GPTQAP} and \citet{Frantar2023SparseGPTML} that their methods are optimized for the largest LLMs. This evident limitation on smaller (yet still substantial) LLMs highlights the imperative for additional performance improvement post-compression beyond just parameter adjustment.


\subsection{Forgotten, or Displaced? A Two-Way Argument}
Investigating whether knowledge in compressed models is forgotten or merely displaced presents a complex challenge. However, discerning between these two scenarios is feasible by examining the nature of intervention required to reinstate the model's performance in downstream tasks.

\begin{itemize}
    \item \underline{Forgetfulness} implies that the compression process irrevocably eliminates certain knowledge. Integrating an external knowledge source becomes essential to recuperate performance, as this process essentially replenishes the lost information.
    
    \item \underline{Displacement} posits that the inherent knowledge within these models is not irrevocably erased but instead shifted internally, leading to the inefficacy of the established inference pathways. In this context, input-side augmentation or instructions are needed to ``redirect" the internal self-attention. This enables the re-engagement of the pre-existing, albeit repositioned, knowledge in the compressed LLM, thereby aiding in the recuperation of its performance.
\end{itemize}

We position LoRA \cite{Hu2021LoRALA} and prompting to correlate respectively with our hypothesis on ``\textbf{knowledge forgotten}" and ``\textbf{knowledge displaced}." LoRA tackles ``forgetfulness" by fundamentally altering the model's structure, specifically the weights in the self-attention and feedforward neural network (FFN) layers, thereby reintegrating knowledge lost due to compression. Prompting, in contrast, operates by subtly influencing the self-attention mechanism without changing the underlying weights, thus redirecting the model's existing but less accessible knowledge. 





\section{Methods and Experiments}

\subsection{From Basic Prompting to IDP}
Building upon the initial work of \citet{Xu2023CompressTP}, which utilized prompting to enhance the performance of compressed models measured by perplexity, we identify a crucial limitation in this approach. As \citet{jaiswal2023compressing} highlights, perplexity alone may not fully capture a model's actual behavior in compression scenarios. To (in)validate this finding, we contrast the perplexity metric with the aggregated accuracy performance across nine downstream tasks using varying prompt lengths, aiming to discern any discrepancies between these two metrics. Our findings, illustrated in Figure \ref{fig:seq_iso}, reveal a notable perplexity-to-performance gap that becomes more pronounced with longer prompts. This divergence not only corroborates the observations of \citet{jaiswal2023compressing} but also underscores the limitations in relying solely on perplexity as a performance indicator, as initially proposed by \citet{Xu2023CompressTP}.

We show that the presumed efficacy of extended prompts is, in fact, compromised by their intrinsic rigidity, leading us to a conclusion: further improvement of prompting hinges on the precise alignment of the right prompt with the appropriate input rather than the elongation of a single prompt. This concept parallels ensemble methods but requires a departure from iterative or training-intensive approaches, such as those found in \citet{Lester2021ThePO} or \citet{peng2023model}, which significantly increase training time and inference costs. To circumvent these drawbacks, we introduce \textit{Inference-time Dynamic Prompting (IDP)}. IDP enables one-shot input-to-prompt matching with minimal latency increase to inference time. This strategy aligns prompts more effectively with inputs and incurs little computational overhead, marking a stepped improvement in prompting for compressed model performance recovery.

\input{figures/gptq_seq}

\subsubsection{IDP Methodology}
\input{figures/prompt_ensemble_figure}

In prompt tuning, we introduce an additional token sequence, termed as \(P\), preceding the input sequence to improve the predicted output likelihood, \( Pr_{\theta}(Y|[P; X]) \), where \( \theta \) are the static parameters. The sequence \(P ={p_1, p_2, ... p_n}\) is defined by its learnable parameters, \( \theta_{p} \in \mathbb{R}^{n\times e} \), with \( n \) being the prompt tokens count and \( e \) as their embedding size.

When we extend to a collection of \( m \) prompts, represented as \( Z = {P_1, P_2, ..., P_m} \), each prompt has distinct trained parameters. Thus, the modified likelihood of \( Y \) becomes \( Pr(Y|[Z; X]) \). Let's consider the layer-wise token attention as \( A \in \mathbb{R}^{b \times h \times tk \times tk} \), where \( tk \) stands for the combined token count of \( Z \) and \( X \). For simplicity, we'll take \( b \) and \( h \) as one.

To facilitate Inference-time Dynamic Prompting, we introduce two modifications to \( A \): \textbf{Firstly}, we prevent interactions among the prompts in \( Z \) by setting their inter-attention, \( A_{[Z_i: Z_j]} \), to \( -\infty \). This constraint is twofold: Individual prompts have distinct training and do not share contextual relevance. Mixing them during inference can alter their inherent definitions, affecting the performance. Additionally, by eliminating inter-prompt attention, we can pre-cache the KV (Key, Value) for the prompts; this enables us to amortize the cost of processing. \textbf{Secondly}, for dynamic prompt selection, we measure the mean attention from input-to-prompt and select the prompt attracting the 
maximum overall input attention: \( \argmax(\{\overline{A}_{[Z_i:X]}| \forall i \in [1,m]) \). In the final phase of the self-attention mechanism, we use an attention mask to discard any unintended prompts, ensuring they do not modify the main input sequence and improve our inference latency. The process is depicted in Figure \ref{fig:prompt_ensemble}.




\subsection{Experimental Comparison: IDP recovers performance better or comparable than LoRA}

\subsubsection{Settings}

\paragraph{Compressed Models} We utilize OPT-6.7b \citep{Zhang2022OPTOP} and Llama-7b \citep{Touvron2023LLaMAOA} as foundational models, both featuring an embedding size “e” of 4096. For compression, we apply GPTQ \cite{Frantar2022GPTQAP} and SparseGPT \cite{Frantar2023SparseGPTML} to achieve 3-bit quantization and 50\% pruning, respectively. In our discussion, we will primarily focus on the quantization approach, as the pruning process exhibits a very similar pattern.

\paragraph{Find-tuning Dataset} We derive two configurations for each compression technique, each optimized on one of the large-scale text datasets: C4 \citep{2020t5} and Wikitext \citep{merity2016pointer}. To maintain a controlled experimental space, our fine-tuning of various baseline techniques is restricted to the identical dataset used initially to calibrate our model compression. We utilize two distinct prompts for IDP-specific settings, "$m$" and "$n$," being 50 and 100.

\paragraph{Validation Tasks} To gauge the genuine comprehensive performance of LLMs, we identify a suite of evaluation tasks that encapsulate three fundamental domains of cognition: world knowledge, common reasoning, and language understanding. Among the many available tasks, we distilled our focus to a curated list of nine that we deemed most representative.

For the domain of world knowledge, our chosen evaluative tasks were ARC-challenge \& ARC-easy \citep{Clark2018ThinkYH}, SCIQ \citep{Welbl2017CrowdsourcingMC}, WebQS \citep{Berant2013SemanticPO}, and TriviaQA \citep{joshi2017triviaqa}. Tapping into the breadth of language understanding benchmarks, we centered our attention on Hellaswag \citep{zellers2019hellaswag}, Lambada \citep{paperno_denis_2016_2630551}, and WinoGrande \citep{sakaguchi2019winogrande}. Lastly, for common reasoning, we identified PIQA \citep{bisk2019piqa} as our touchstone. Notably, all the tasks we adopted are structured in a multiple-choice format. 

\paragraph{Baseline Methods} We gravitated toward three methodologies to test our hypothesis. We use prompt-tuning \citep{Lester2021ThePO}, prefix-tuning \citep{Li2021PrefixTuningOC}, and LoRA \citep{Hu2021LoRALA} as our representative candidates. For consistent benchmarks across these techniques, we establish the following criteria: 1) The aggregate count of training tokens is standardized at 40,960,000 tokens. Our decision on the total token count draws inspiration from \cite{Xu2023CompressTP}. 2) In alignment with \cite{Frantar2022GPTQAP}, we adopt AdamW as our optimization algorithm. Our chosen learning rate stands at 2e-4 with a weight decay set at 1e-5. All three methods are then fine-tuned using compressed LLM following the described settings with LLama-7b and OPT-6.7b.

\input{tables/gptq_results}

\input{tables/sparse_results}

\subsubsection{Results}

We compare several post-compression performance recovery techniques with IDP. We report our findings in Table \ref{tab:gptq} and Table \ref{tab:sparsegpt}. From these tables, we draw the following insights:

\begin{itemize}

    \item \textbf{Performance Recovery:} Overall, our testing shows that most techniques, IDP included, modestly improve performance in both quantization and pruning scenarios. The sole outlier is “ptune” or prefix-tuning, which, in contrast, reduced the performance of our models across all tasks. We also noted that quantization provides a better chance for performance recovery compared to pruning. Comparing baseline performances with the top performers, GPTQ shows an average improvement of about 1\%, compared to  SparseGPT's modest increase of 0.37\%. This difference likely stems from pruning's parameter removal, which has stricter limits on performance enhancement. However, IDP stands out in both scenarios, achieving the higher-than-average performance improvement. 
    
    \item \textbf{Knowledge Domain Adaptation:} By dividing our tasks into world knowledge, common reasoning, and language understanding categories, we draw insights on ``knowledge displaced" and ``knowledge forgotten." We observed that redirection methods such as IDP are more effective for world knowledge tasks than integrated methods like LoRA. This suggests that in cases of factual knowledge, simple input redirection can effectively bring back the lost information in compressed models. Conversely, tasks that demand nuanced understanding, like those involving language comprehension, benefit more from the added parameters and external knowledge sources provided by techniques like LoRA. Nevertheless, in scenarios where IDP lacks, the difference in performance is marginal -- less than 0.2\%.
    
\end{itemize}

\subsection{IDP is remarkably more efficient}

 We examined the link between the method's parameter size and performance, as detailed in Figure \ref{fig:param_perf}. Our findings show that IDP is much more efficient than LoRA for compression recovery. For example, when fine-tuning the Llama-7b model with QPTQ settings, LoRA's parameters range between 4.4 to 8.9 million, while IDP uses only around 0.8 million, leading to substantial space savings of 81\% to 91\% — a notable \textbf{20-fold} reduction. Additionally, prompting tends to have a faster inference speed. Basic inference testing shows prompting incurs at most 0.37s versus LoRA's 0.62s for an input batch of 16 and a sequence length of 1024 -- this is a substantial \textbf{60\% improvement in speed}. Despite the smaller size, IDP generally sees a modest average improvement of 1\% across the nine tasks evaluated. For further details on the performance and parameter size, refer to Table \ref{tab:gptq} and Figure \ref{fig:param_perf}, and our appendix provides a detailed explanation of how the total number of parameters was calculated for both LoRA and IDP.

Finally, we underscore the robustness of IDP's performance, irrespective of prompt in Figure \ref{fig:robust}. This figure reveals a variance of less than 1\% in average accuracy performance, yet with a 5-fold reduction in token size. Notably, even with a modest average of 20 tokens, IDP adeptly facilitates performance recovery, surpassing the compressed baseline. This evidence positions IDP as not only efficient in parameter utilization but also as a resilient mechanism for enhancing performance in the wake of model compression.

\input{figures/peft_gptq_perf}


\input{figures/cosim}

\section{More Studies}
\subsection{Evaluating Knowledge Forgetfulness and Displacement }

We employed a detailed visualization of the layer-wise attention and activation matrices to validate our hypothesis. Opting for cosine similarity over magnitude differences as our analytical tool, we aim to understand the distribution differences rather than magnitude. Our findings are  presented in Figures \ref{fig:cos}, and \ref{fig:robust}, leading to several key observations:

\input{figures/IDP_robustness}

    $\circled{1}$  When compared to LoRA, the attention mechanism of both prompting/IDP markedly diverges from the baseline, hinting at a potential contextual redirection. Conversely, the activation patterns echo similarities with LoRA. Given that LoRA incorporates a residual network at every layer to maintain congruity and prompting only at the self-attention, this semblance is unexpected.
    
    $\circled{2}$ These observations imply that prompting/IDP can tap into latent knowledge within the model. This is further supported by the data in Table \ref{tab:gptq} and Table \ref{tab:sparsegpt}, which show a propensity of prompting/IDP for tasks involving world knowledge. These tasks rely on the model's internal knowledge base, reinforcing our conclusion about the efficacy of prompting/IDP in accessing embedded information.

    $\circled{3}$ Additionally, IDP demonstrates remarkable consistency in information retrieval. As evidenced in Figure \ref{fig:robust}, it maintains stable performance across a range of prompt sizes. This suggests that even with fewer tokens, knowledge rerouting via IDP remains effective, opening avenues for future optimizations and refinements in its application.
    
    $\circled{4}$ Finally, our analysis of prefix-tuning indicates its tendency to align with the original attention patterns of the model. However, as shown in Figure \ref{fig:cos}, its activation patterns significantly deviate, hinting at a potential shortfall in redirecting knowledge.

These insights strongly endorse the notion of ``redirection" as the more effective mechanism for recovering performance in compressed models. 

\subsection{More ablation studies}
\input{tables/gptq_prompt_ensemble}

\input{figures/ensemble-polar}

We used IDP strategy with two distinct prompts of differing lengths, both trained using the same dataset to streamline our experimental parameters. We subsequently evaluated against our task benchmark, with the comprehensive findings cataloged in Table \ref{tab:ensemble}. In a complementary visual aid, Figure \ref{fig:ensemble-polar} highlights the percentage differences in performance against the baseline quantized models, providing an at-a-glance understanding of the performance gains across individual tasks.


Our analysis showed that IDP subtly enhances average accuracy. This is evident in our results with OPT and Llama models, where IDP showed a modest improvement of $0.5\%$ and $0.42\%$, respectively. This contrasted with the outcomes of basic prompt concatenation, which yielded only a $0.16\%$ increase and even a decrease of $-1.03\%$. While these findings, detailed in Table \ref{tab:ensemble}, might not be groundbreaking, they highlight the potential of zero-shot input-to-prompt matching for compression recovery for various knowledge domains.


Further, in our examination of quantized foundation models, as shown in Figure \ref{fig:ensemble-polar}, we noted areas where IDP demonstrated a slight but consistent superiority. Specifically, OPT models showed this incremental benefit in tasks such as Sciq, Triviqa, and Webqs, all falling within the world knowledge domain. Similarly, the Llama models exhibited slight improvements in tasks like Webqs, Arc, and Winogrand, with gains ranging between 1\%-1.5\%. 


\section{Conclusion}


In this study, we focused on understanding the impact of compression on LLMs and explore ways to mitigate its negative effects. We explore two key hypotheses: knowledge forgotten and knowledge displaced by analyzing the effectiveness of parameter-efficient tuning method like LoRA and prompting. A highlight from our study is the introduction Inference-time Dynamic Prompting , a light-weight approach to enhance traditional prompting. Empirically, we find IDP-based prompting perform on-par or better than relearning approach like LoRA while being significantly smaller in size with faster inference speed. Additionally, our visualization of the intermediate embeddings within LLMs suggests redirection through instruction is a more beneficial way to regain similar activation output. Collectively, our findings advocate the hypothesis of ``knowledge displaced" as the critical factor behind performance decline post-compression, providing valuable insights into the mechanisms at play and paving the way for more efficient recovery strategies.



\section{Impact Statements}
 This paper presents work whose goal is to advance the field of Machine Learning. There are many potential societal consequences of our work, none which we feel must be specifically highlighted here.
 
\bibliography{icml2024}
\bibliographystyle{icml2024}
\newpage
\appendix
\onecolumn

\section{Number of Parameters}
\paragraph{LoRA} In the LoRA method \cite{Hu2021LoRALA}, the total number of parameters is calculated as the sum of all low-rank projected layers within the LLMs. Consider the output of a singular LoRA layer as $x_{lora} = BAx$, where $B \in \mathbb{R}^{d \times r}$ and $A \in \mathbb{R}^{r \times d}$ represent the learnable low-rank decomposition matrices. Here, $d$ denotes the token's intermediate activation dimension, and $r$ is the number of low-rank dimensions. For any given layer, the total parameter count is computed as $8d_{act}r + 4d_{inter}r$, accounting for Key, Value, Query, and Output components in the self-attention mechanism, as well as the two Multi-Layer Perceptrons (MLPs) in the FeedForward Network (FFN). In our specific configuration, we set $r$ to 2, 3, and 4, with $d_{act}$ as 4096 and $d_{inter}$ as 11008.

\paragraph{Prompting / IDP} In IDP, the total number of parameters is calculated by multiplying the total number of tokens, denoted as $t$, by the dimension of the token's embeddings, $d$. In our configuration, $t$ is set to 150 and $d$ equals 4096. For prompting, the values of $t$ are 26, 50, and 100, corresponding to different settings.

\paragraph{Prefix-tuning} For Prefix-tuning (Ptune), the total number of parameters can be determined using the formula $L * t * d$. Here, $L$ represents the number of layers, $t$ the number of tokens, and $d$ the dimension of the token embeddings. $L$ is 32, $d$ equals 4096, and the values of $t$ are 26, 50, and 100.

\end{document}

%% file: figures/compression_baseline.tex
\begin{figure}[htbp]
\centering

\includegraphics[width=\linewidth]{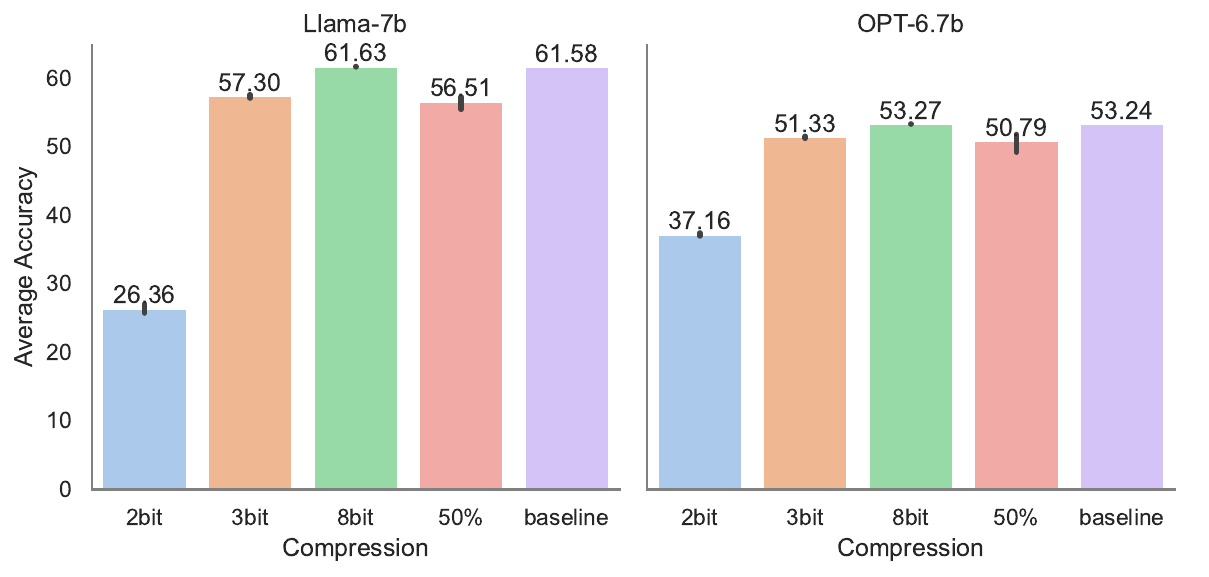} 

\caption{This figure presents a comparative analysis of the performance of compressed models using GPTQ for quantization and SparseGPT for pruning. The models were compressed leveraging either C4 or Wikitext datasets. Their average performance is depicted across a spectrum of nine tasks, each representing diverse knowledge domains.}
\label{fig:compression_baseline}
\end{figure}

%% file: figures/gptq_seq.tex
\begin{figure}[htbp]
\centering
\includegraphics[width=\linewidth]{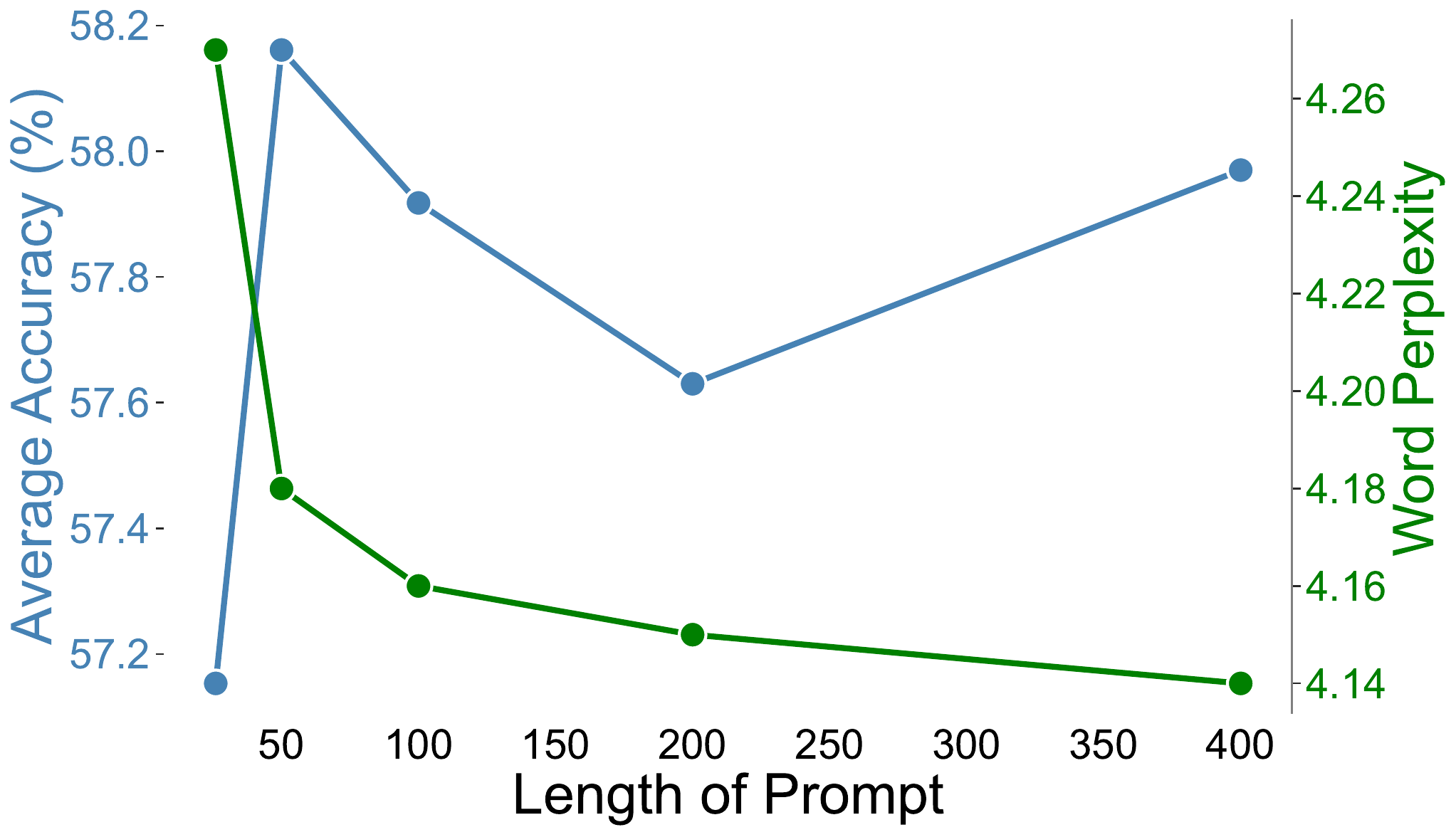} 

\caption{Using a 3-bit quantized Llama-7b model fine-tuned on C4 dataset, we contrast the average accuracy across nine tasks against its word's perplexity score across various prompt lengths. A longer sequence length improves perplexity but does not always sustain better performance.}
\label{fig:seq_iso}
\end{figure}

%% file: figures/prompt_ensemble_figure.tex
\begin{figure*}[htbp]
\centering
\includegraphics[width=\linewidth]{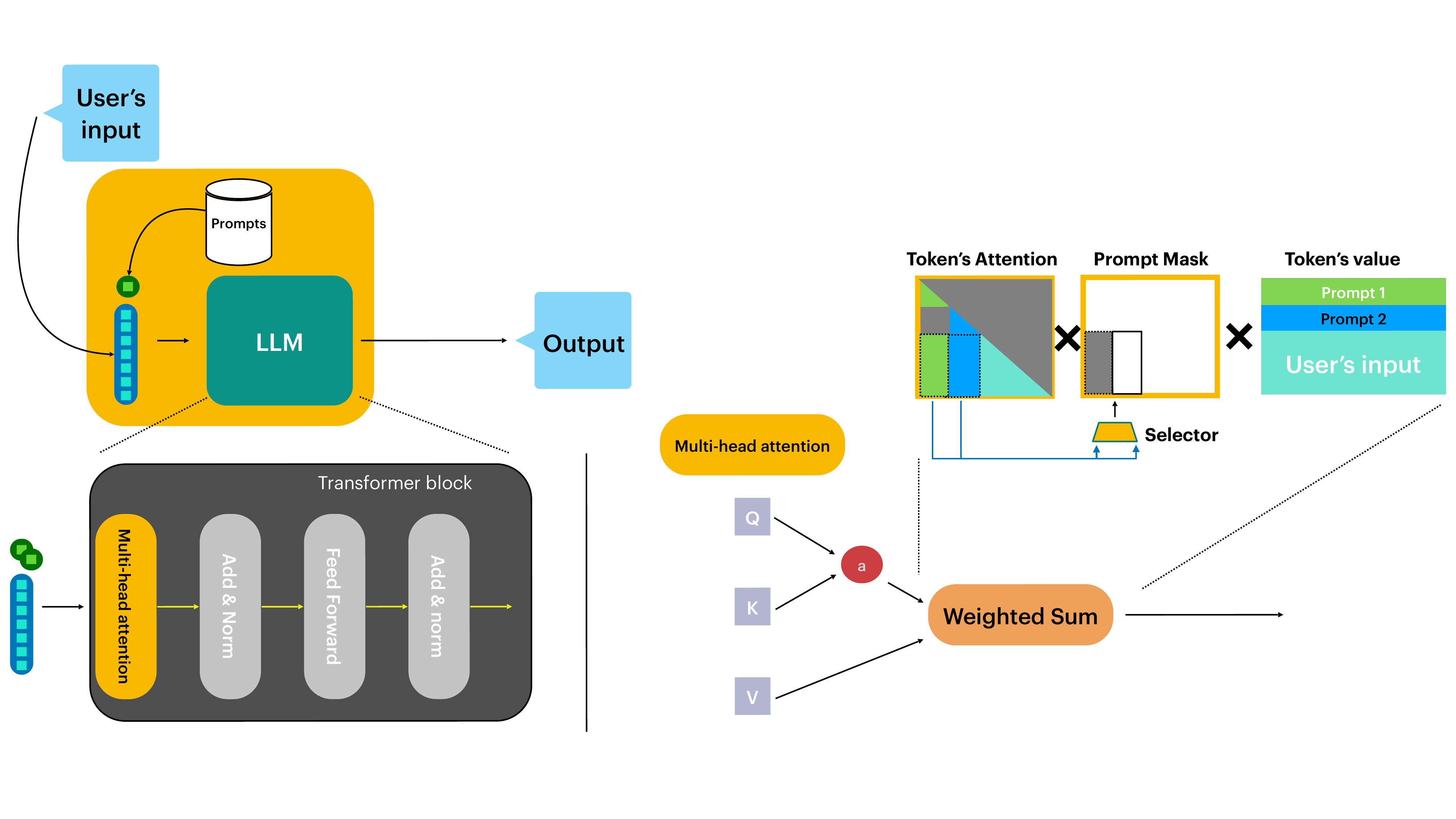} 
\vspace{-20pt}
\caption{This figure underscores the key advantage of inference-time dynamic prompting (IDP): its minimalistic yet effective design. By making straightforward alterations to the existing weighted sum operation and using the existing attention matrix for prompt selection, IDP accomplishes its objectives without incurring any additional parameter costs. }
\label{fig:prompt_ensemble}
\end{figure*}

%% file: tables/gptq_results.tex
\begin{table*}
\caption{This table summarizes the results for 3-bit GPTQ across all nine tasks for multiple fine-tuning baselines and our IDP. World, Common, and Language are performance averages across tasks within those knowledge domains. Average is the average performance across all nine tasks.}
\begin{adjustbox}{width=\linewidth,center}

\begin{tabular}{lll|rrrrr|c||r|c||rrr|c|||c}
\toprule
Model & Type & Param &  arcE & arcC & sciq & webqs & triviaqa & \textbf{World} & piqa & \textbf{Common} & hellaswag & lambada & winogrande & \textbf{Language} & \textbf{Average} \\
\midrule
Llama-7b & --- & --- & 71.46 & 37.71 & 92.60 & 17.96 & 33.02 & 50.55 & 76.01 & 76.01 & 53.11 & 68.58 & 67.48 & 63.06 & 57.55 \\
Llama-7b & lora & 4.4M &70.08 & 37.12 & \cellcolor{green}93.50 & 17.67 & 34.11 & 50.50 & 77.04 & 77.04 & 54.47 & \cellcolor{green}70.48 & 67.40 & \cellcolor{green}\textbf{64.12} & 57.99 \\
Llama-7b & lora & 6.7M &71.09 & 36.69 & 93.00 & 17.47 & 34.73 & 50.60 & 76.44 & 76.44 & \cellcolor{green}54.55 & 70.23 & 67.09 & 63.96 & 57.92 \\
Llama-7b & lora & 8.9M & 70.62 & 37.12 & 93.30 & 17.86 & 34.86 & 50.75 & 76.77 & \cellcolor{green}\textbf{76.77} & 54.27 & 70.33 & 67.40 & 64.00 & 58.06 \\
Llama-7b & prompt & 0.1M &  71.97 & 38.40 & 92.90 & 20.47 & 33.20 & 51.39 & 75.84 & 75.84 & 53.75 & 69.45 & 67.17 & 63.46 & 58.13 \\
Llama-7b & prompt & 0.2M &  71.51 & 38.31 & 92.10 & \cellcolor{green}21.11 & 34.56 & 51.52 & 75.84 & 75.84 & 53.92 & 69.69 & \cellcolor{green}68.75 & 64.12 & 58.42 \\
Llama-7b & prompt & 0.4M &  72.01 & 39.16 & 91.80 & 21.60 & 34.43 & 51.80 & 75.95 & 75.95 & 54.33 & 69.49 & 67.01 & 63.61 & 58.42 \\
Llama-7b & ptune & 3.1M & 70.24 & 36.77 & 91.40 & 14.42 & 30.42 & 48.65 & 75.73 & 75.73 & 53.40 & 66.49 & 63.77 & 61.22 & 55.85 \\
Llama-7b & ptune & 6.5M & 69.57 & 34.81 & 91.30 & 15.55 & 30.65 & 48.38 & 75.30 & 75.30 & 52.98 & 64.84 & 63.22 & 60.35 & 55.36 \\
Llama-7b & ptune & 13.1M &  69.32 & 34.73 & 88.70 & 16.14 & 27.84 & 47.35 & 74.59 & 74.59 & 52.01 & 64.35 & 64.17 & 60.18 & 54.65 \\
\hdashline

Llama-7b & IDP & 0.6M &  \cellcolor{green}72.43 & \cellcolor{green}39.76 & 92.50 & 19.83 & \cellcolor{green}36.39 & \cellcolor{green}\textbf{52.18} & 76.44 & 76.44 & 53.96 & 70.25 & 67.56 & 63.92 & \cellcolor{green}\textbf{58.79} \\
\midrule
OPT-6.7b & --- &  --- & 64.77 & 29.01 & 89.40 & 9.50 & 17.90 & 42.12 & 75.24 & 75.24 & 48.57 & 65.34 & 63.54 & 59.15 & 51.47 \\
OPT-6.7b & lora & 4.7M &  63.55 & 28.75 & 88.50 & 11.42 & 18.84 & 42.21 & 76.22 & \cellcolor{green}\textbf{76.22} & \cellcolor{green}49.14 & 66.16 & 63.46 & 59.59 & 51.78 \\
OPT-6.7b & lora & 7.1M &  64.27 & 29.01 & 89.20 & 11.07 & 18.95 & 42.50 & 75.90 & 75.90 & 48.89 & 66.50 & \cellcolor{green}64.40 & \cellcolor{green}\textbf{59.93} & 52.02 \\
OPT-6.7b & lora & 9.4M &  64.06 & \cellcolor{green}29.35 & 88.20 & 13.24 & 18.90 & 42.75 & 76.01 & 76.01 & 49.12 & 66.64 & 63.93 & 59.90 & 52.16 \\
OPT-6.7b & prompt & 0.1M &  \cellcolor{green}64.27 & 28.41 & 89.80 & 10.73 & 18.22 & 42.50 & 76.01 & 76.01 & 49.05 & 65.34 & 63.22 & 59.20 & 51.79 \\
OPT-6.7b & prompt & 0.2M &  64.94 & 28.84 & 89.90 & 10.88 & 18.80 & 42.67 & 75.63 & 75.63 & 49.13 & 65.96 & 63.77 & 59.62 & 51.98 \\
OPT-6.7b & prompt & 0.4M& 64.60 & 28.50 & 89.70 & 11.52 & 18.76 & 42.62 & 76.12 & 76.12 & 48.82 & 65.90 & 63.54 & 59.42 & 51.94 \\
OPT-6.7b & ptune & 3.1M&  63.05 & 28.84 & 89.00 & 10.73 & 18.39 & 42.00 & 75.95 & 75.95 & 48.38 & 64.68 & 60.85 & 57.97 & 51.10 \\
OPT-6.7b & ptune & 6.5M&  62.88 & 28.58 & 88.80 & 10.43 & 18.34 & 41.81 & 75.79 & 75.79 & 48.54 & 65.17 & 60.93 & 58.21 & 51.05 \\
OPT-6.7b & ptune & 13.1M & 62.54 & 29.18 & 88.60 & 10.43 & 18.37 & 41.82 & 75.52 & 75.52 & 48.72 & 65.32 & 63.38 & 59.14 & 51.34 \\
\hdashline

OPT-6.7b & IDP &  0.6M  & 64.18 & 28.67 & \cellcolor{green}90.40 & \cellcolor{green}11.96 & \cellcolor{green}19.05 & \cellcolor{green}\textbf{42.85} & 76.17 & 76.17 & 49.03 & \cellcolor{green}66.82 & 63.22 & 59.69 & \cellcolor{green}\textbf{52.17} \\
\bottomrule
\end{tabular}
\end{adjustbox}
\label{tab:gptq}
\end{table*}

%% file: tables/sparse_results.tex
\begin{table*}
\caption{This table summarizes the results for 50\% unstructured sprase using SparseGPT across all nine tasks for multiple fine-tuning baselines and our IDP. World, Common, and Language are performance averages across tasks within those knowledge domains. Average is the average performance across all nine tasks.}
\begin{adjustbox}{width=\linewidth,center}

\begin{tabular}{lll|rrrrr|c||r|c||rrr|c|||c}
\toprule
Model & Type & Param &  arcE & arcC & sciq & webqs & triviaqa & \textbf{World} & piqa & \textbf{Common} & hellaswag & lambada & winogrande & \textbf{Language} & \textbf{Average} \\
\midrule
Llama-7b &--- & --- & 70.33 & 37.03 & \cellcolor{green}93.50 & 14.07 & 28.88 & 48.76 & 77.04 & 77.04 & 51.68 & \cellcolor{green}74.54 & \cellcolor{green}68.03 & \cellcolor{green}64.75 & 57.23 \\
Llama-7b & lora & 4.4M & 71.04 & 37.63 & 91.90 & 14.47 & \cellcolor{green}33.28 & 49.66 & 76.99 & 76.99 & 53.98 & 70.95 & 67.17 & 64.03 & 57.49 \\
Llama-7b & lora &  6.7M & 70.79 & 36.69 & 92.40 & 15.85 & 33.02 & 49.75 & 76.71 & 76.71 & 53.91 & 71.03 & 68.03 & 64.32 & 57.60 \\
Llama-7b & lora & 8.9M & 71.04 & 37.88 & 92.10 & \cellcolor{green}14.86 & 32.85 & 49.75 & 77.20 &\cellcolor{green} 77.20 & \cellcolor{green} 54.01 & 70.70 & 68.03 & 64.25 & \cellcolor{green}57.63 \\
Llama-7b & prompt &  0.1M& 71.59 & 38.74 & 93.10 & 15.21 & 29.66 & 49.66 & 77.04 & 77.04 & 53.48 & 71.24 & 67.48 & 64.07 & 57.50 \\
Llama-7b & prompt & 0.2M & 71.38 & 38.57 & 92.20 & 14.86 & 30.48 & 49.50 & 77.15 & 77.15 & 53.75 & 71.76 & 67.09 & 64.20 & 57.47 \\
Llama-7b & prompt & 0.4M & 71.38 & 38.31 & 92.60 & 14.86 & 30.86 & 49.60 & 77.31 & 77.31 & 53.97 & 70.99 & 67.17 & 64.04 & 57.49 \\
Llama-7b & ptune & 3.1M & 63.17 & 32.59 & 88.20 & 11.81 & 24.60 & 44.07 & 72.63 & 72.63 & 50.18 & 64.97 & 56.91 & 57.35 & 51.67 \\
Llama-7b & ptune & 6.5M & 67.17 & 34.90 & 88.70 & 12.11 & 24.74 & 45.52 & 74.76 & 74.76 & 50.36 & 65.59 & 59.12 & 58.36 & 53.05 \\
Llama-7b & ptune & 13.1M & 65.78 & 31.40 & 87.20 & 11.61 & 21.97 & 43.59 & 74.21 & 74.21 & 49.77 & 63.87 & 59.43 & 57.69 & 51.69 \\
\hdashline
Llama-7b & IDP & 0.6M  & \cellcolor{green}72.05 & \cellcolor{green}39.08 & 92.90 & 14.91 & 30.35 & \cellcolor{green}49.86 & 77.09 & 77.09 & 53.90 & 70.35 & 67.17 & 63.81 & 57.53 \\
\midrule
\midrule
OPT-6.7b &---& --- & 63.01 & 28.41 & 89.40 & 9.69 & 17.79 & 41.66 & 75.19 & 75.19 & 47.67 & \cellcolor{green}70.56 & 63.93 & \cellcolor{green} 60.72 & 51.74 \\
OPT-6.7b & lora & 4.7M & \cellcolor{green}64.06 & 29.61 & 88.60 & 10.58 & 18.26 & 42.22 & 75.57 & 75.57 & 48.52 & 66.60 & 64.33 & 59.82 & 51.79 \\
OPT-6.7b & lora & 7.1M & 63.93 & 29.78 & 88.20 & 10.14 & 18.48 & 42.11 & 75.90 & \cellcolor{green}75.90 & 48.58 & 66.45 & \cellcolor{green}64.56 & 59.86 & 51.78 \\
OPT-6.7b & lora & 9.4M & 62.84 & \cellcolor{green}29.86 & 88.30 & 10.33 & 18.79 & 42.02 & 75.41 & 75.41 & \cellcolor{green}48.76 & 66.49 & 65.19 & 60.15 & 51.77 \\
OPT-6.7b & prompt & 0.1M & 63.09 & 28.58 & \cellcolor{green}90.70 & 12.30 & 18.75 & 42.68 & 75.14 & 75.14 & 48.40 & 68.78 & 63.69 & 60.29 & 52.16 \\
OPT-6.7b & prompt & 0.2M & 63.68 & 29.44 & 90.60 & 12.40 & 18.36 & 42.90 & 75.24 & 75.24 & 48.58 & 67.86 & 63.22 & 59.89 & 52.15 \\
OPT-6.7b & prompt & 0.4M & 64.06 & 29.27 & 89.60 & 12.80 & 19.12 & 42.97 & 75.19 & 75.19 & 48.49 & 67.49 & 63.61 & 59.86 & 52.18 \\
OPT-6.7b & ptune & 3.1M & 61.03 & 28.50 & 86.90 & 13.09 & 19.46 & 41.80 & 72.74 & 72.74 & 46.44 & 62.08 & 59.67 & 56.06 & 49.99 \\
OPT-6.7b & ptune & 6.5M & 63.01 & \cellcolor{green}29.86 & 88.00 & 9.40 & 17.10 & 41.47 & 75.08 & 75.08 & 47.84 & 64.89 & 61.80 & 58.18 & 50.78 \\
OPT-6.7b & ptune & 13.1M & 60.94 & 29.10 & 88.60 & \cellcolor{green}13.53 & \cellcolor{green}19.95 & 42.42 & 73.39 & 73.39 & 46.93 & 62.68 & 62.19 & 57.27 & 50.81 \\
\hdashline
OPT-6.7b & IDP & 0.6M & \cellcolor{green}64.06 & 29.27 & 89.60 & 12.80 & 19.12 & \cellcolor{green}42.97 & 75.19 & 75.19 & 48.49 & 67.49 & 63.61 & 59.86 & \cellcolor{green}52.18 \\
\bottomrule
\end{tabular}
\end{adjustbox}
\label{tab:sparsegpt}
\end{table*}

%% file: figures/peft_gptq_perf.tex
\begin{figure}[htbp]
\centering
\includegraphics[width=\linewidth]{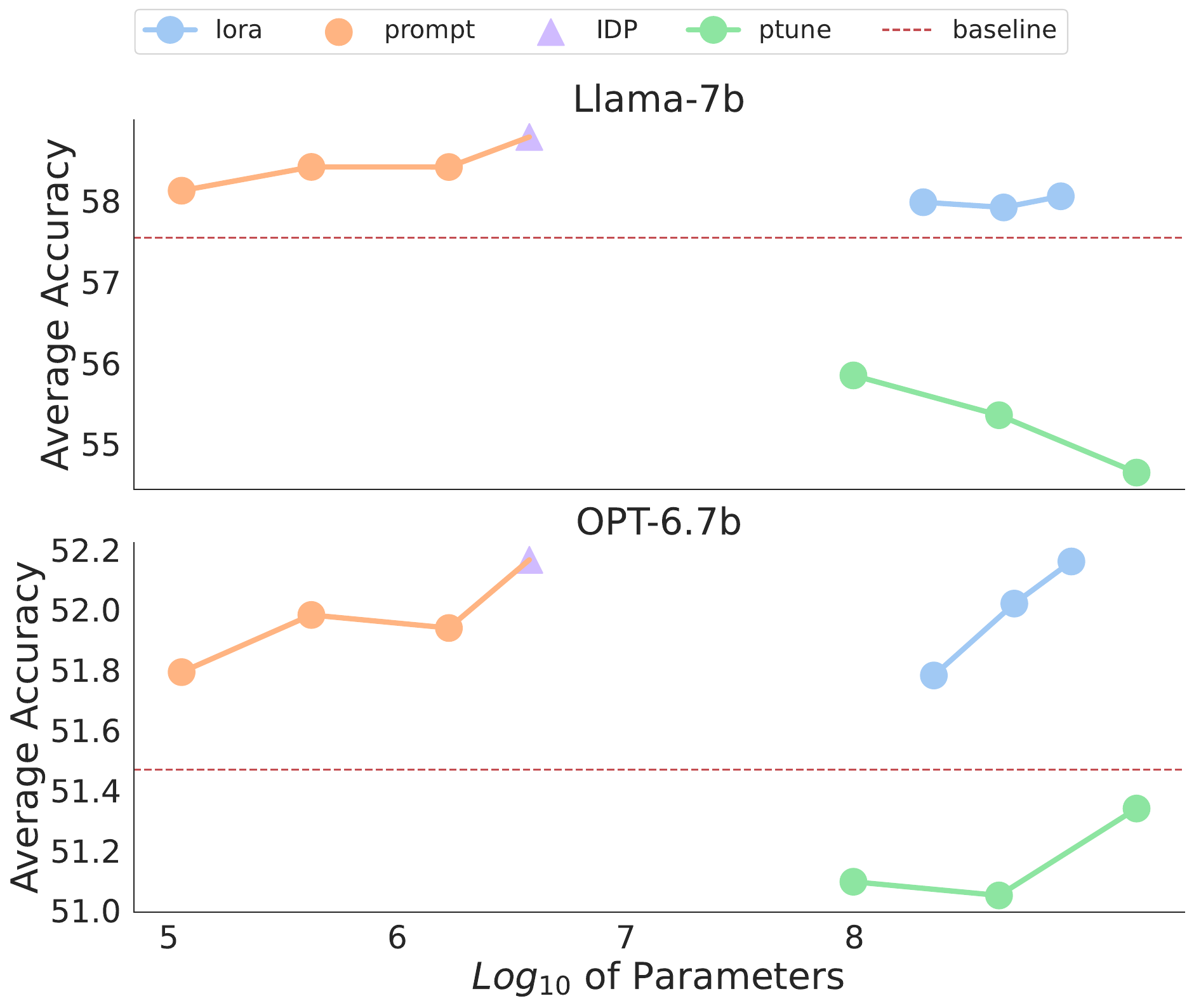} 
\caption{GPTQ LLama-7b/OPT-6.7b average accuracy across nine tasks vs. number of trainable parameters. IDP shows remarkable efficiency and performance comparing to methods parameter-intensive method like LoRA.}

\label{fig:param_perf}
\end{figure}

%% file: figures/cosim.tex
\begin{figure*}
\centering
\includegraphics[width=\linewidth]{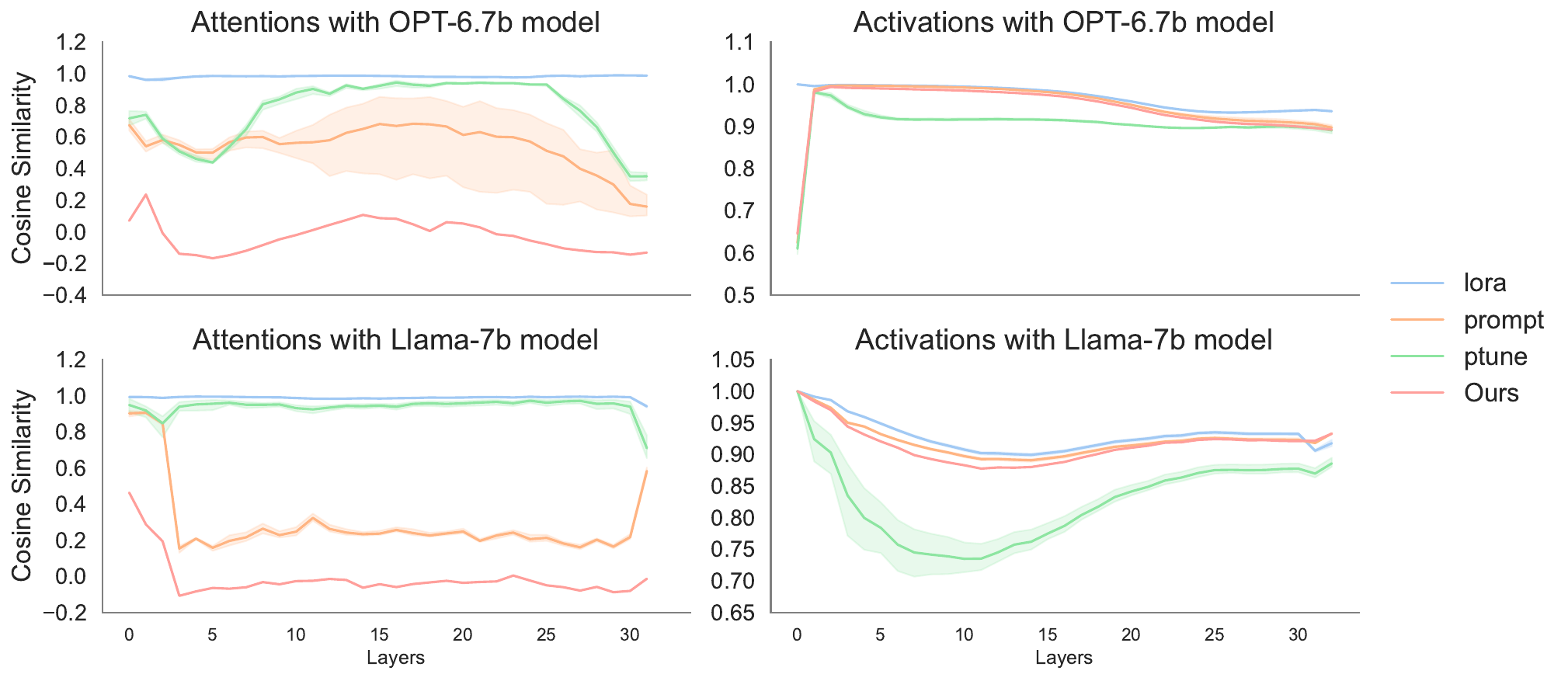} 
\caption{Cosine similarity compares the self-attention and token activation at each layer to an uncompressed baseline using different fine-tuning techniques. A higher cosine score means it's closer to the baseline.}
\label{fig:cos}
\end{figure*}

%% file: figures/IDP_robustness.tex
\begin{figure}[htbp]
\centering
\includegraphics[width=0.8\linewidth]{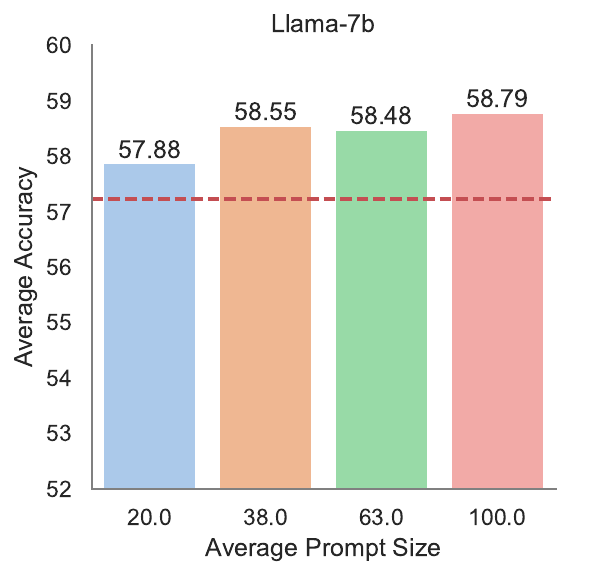} 
\caption{This figure illustrates the average performance over nine tasks using IDP. Results show IDP maintains relatively stable performance working with various average prompt sizes.}
\label{fig:robust}

\end{figure}

%% file: tables/gptq_prompt_ensemble.tex
\begin{table*}
\vspace{-10pt}
\caption{This table includes results for our Inference-time Dynamic Prompting strategy. To illustrate its effectiveness, we also include the results of the individual prompts used along with naive soft-prompts concatenation. 26 and 100 refers to the number of tokens in our prompts.}
\begin{adjustbox}{width=\textwidth,center}
\begin{tabular}{l|rrrrr|c||r|c||rrr|c|||c}
\toprule
Model  & arcE & arcC & sciq & webqs & triviaqa & \textbf{World} & piqa & \textbf{Common} & hellaswag & lambada & winogrande & \textbf{Language} & \textbf{Average} \\
\toprule
OPT-6.7b/26 &  64.94 & 28.84 & 89.90 & 10.88 & 18.80 & 42.67 & 75.63 & 75.63 & 49.13 & 65.96 & 63.77 & 59.62 & 51.98 \\
OPT-6.7b/100 & 64.02 & 27.90 & 89.50 & 11.32 & 18.37 & 42.22 & 76.39 & \cellcolor{green}\textbf{76.39} & 48.81 & 65.42 & 63.22 & 59.15 & 51.66 \\
OPT-6.7b/Concat &  63.80 & 28.50 & 89.40 & 12.30 & 19.55 & 42.71 & 75.79 & 75.79 & 48.92 & 64.72 & 63.85 & 59.16 & 51.87 \\
OPT-6.7b/IDP &  64.18 & 28.67 & 90.40 & 11.96 & 19.05 & \cellcolor{green}\textbf{42.85} & 76.17 & 76.17 & 49.03 & 66.82 & 63.22 & \cellcolor{green}\textbf{59.69} & \cellcolor{green}\textbf{52.17} \\
\midrule

Llama-7b/26 &  71.97 & 38.40 & 92.90 & 20.47 & 33.20 & 51.39 & 75.84 & 75.84 & 53.75 & 69.45 & 67.17 & 63.46 & 58.13 \\
Llama-7b/100 &  71.51 & 38.31 & 92.10 & 21.11 & 34.56 & 51.52 & 75.84 & 75.84 & 53.92 & 69.69 & 68.75 & 64.12 & 58.42 \\
Llama-7b/Concat &  71.17 & 37.80 & 92.30 & 16.88 & 33.84 & 50.40 & 74.92 & 74.92 & 53.34 & 67.18 & 66.46 & 62.33 & 57.10 \\
Llama-7b/IDP &  71.63 & 38.65 & 92.60 & 21.60 & 33.84 & \cellcolor{green}\textbf{51.66} & 76.01 & \cellcolor{green}\textbf{76.01} & 53.97 & 69.67 & 68.98 & \cellcolor{green}\textbf{64.21} & \cellcolor{green}\textbf{58.55} \\
\bottomrule

\end{tabular}
\end{adjustbox}
\label{tab:ensemble}
\end{table*}

%% file: figures/ensemble-polar.tex
\begin{figure*}
\centering
\vspace{-10pt}
\includegraphics[width=0.75\linewidth]{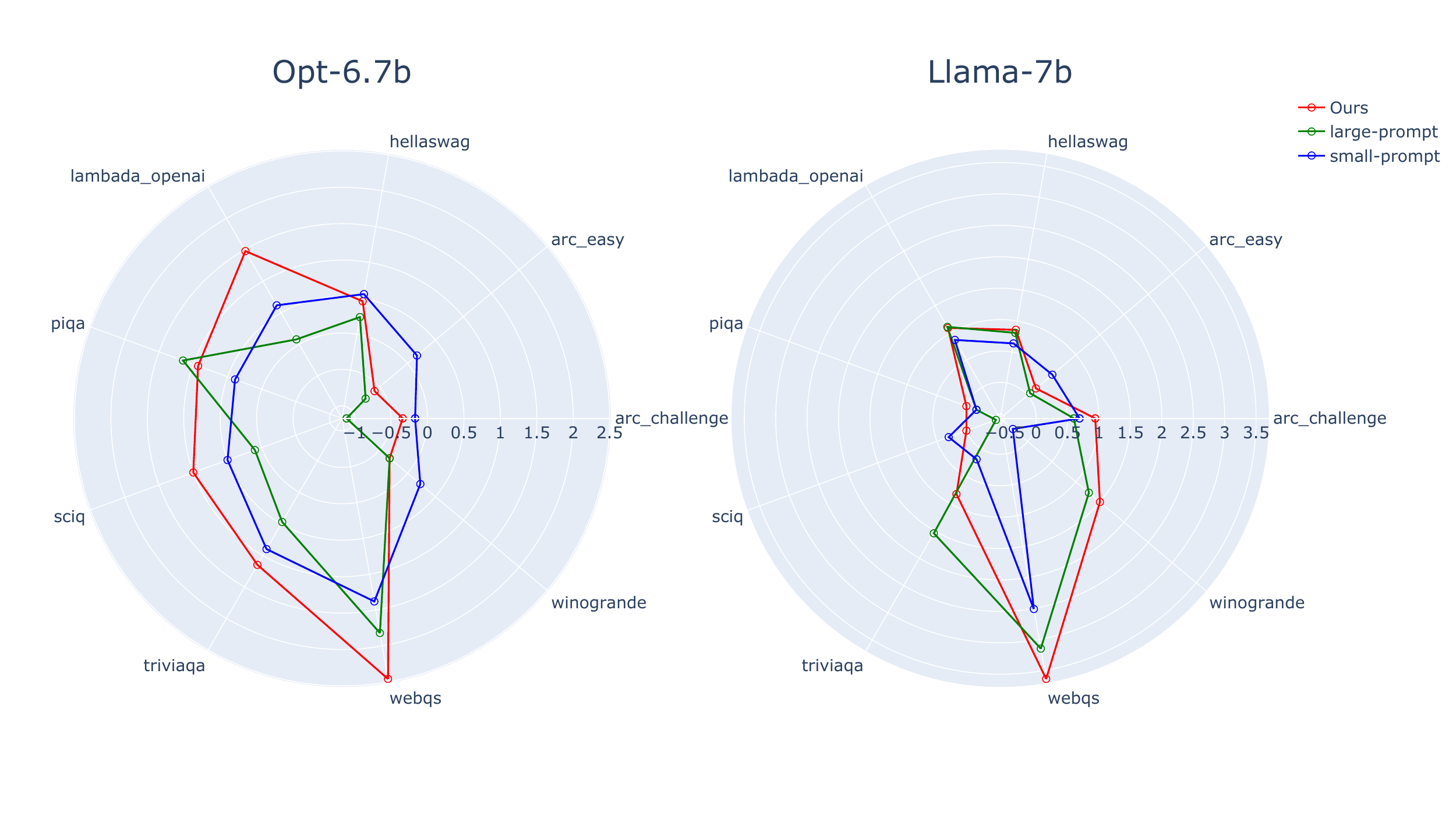} 
\vspace{-40pt}
\caption{This graph shows the percentage performance improvement using two prompts at various lengths compared to a 3-bit quantized baseline for the OPT and LLama models. We've also showcased results from our IDP method, which selects prompts dynamically using the same two prompts. Small and Large correspond to 26 and 100 tokens respectively. }
\label{fig:ensemble-polar}
\vspace{-10pt}
\end{figure*}

%% file: icml2024.bbl
\begin{thebibliography}{27}
\providecommand{\natexlab}[1]{#1}
\providecommand{\url}[1]{\texttt{#1}}
\expandafter\ifx\csname urlstyle\endcsname\relax
  \providecommand{\doi}[1]{doi: #1}\else
  \providecommand{\doi}{doi: \begingroup \urlstyle{rm}\Url}\fi

\bibitem[Berant et~al.(2013)Berant, Chou, Frostig, and Liang]{Berant2013SemanticPO}
Berant, J., Chou, A.~K., Frostig, R., and Liang, P.
\newblock Semantic parsing on freebase from question-answer pairs.
\newblock In \emph{Conference on Empirical Methods in Natural Language Processing}, 2013.
\newblock URL \url{https://api.semanticscholar.org/CorpusID:6401679}.

\bibitem[Bisk et~al.(2019)Bisk, Zellers, Bras, Gao, and Choi]{bisk2019piqa}
Bisk, Y., Zellers, R., Bras, R.~L., Gao, J., and Choi, Y.
\newblock Piqa: Reasoning about physical commonsense in natural language, 2019.

\bibitem[Chen et~al.(2023)Chen, Zaharia, and Zou]{Chen2023FrugalGPTHT}
Chen, L., Zaharia, M.~A., and Zou, J.~Y.
\newblock Frugalgpt: How to use large language models while reducing cost and improving performance.
\newblock \emph{ArXiv}, abs/2305.05176, 2023.
\newblock URL \url{https://api.semanticscholar.org/CorpusID:258564349}.

\bibitem[Clark et~al.(2018)Clark, Cowhey, Etzioni, Khot, Sabharwal, Schoenick, and Tafjord]{Clark2018ThinkYH}
Clark, P., Cowhey, I., Etzioni, O., Khot, T., Sabharwal, A., Schoenick, C., and Tafjord, O.
\newblock Think you have solved question answering? try arc, the ai2 reasoning challenge.
\newblock \emph{ArXiv}, abs/1803.05457, 2018.
\newblock URL \url{https://api.semanticscholar.org/CorpusID:3922816}.

\bibitem[Frantar \& Alistarh(2023)Frantar and Alistarh]{Frantar2023SparseGPTML}
Frantar, E. and Alistarh, D.
\newblock Sparsegpt: Massive language models can be accurately pruned in one-shot.
\newblock \emph{ArXiv}, abs/2301.00774, 2023.
\newblock URL \url{https://api.semanticscholar.org/CorpusID:255372747}.

\bibitem[Frantar et~al.(2022)Frantar, Ashkboos, Hoefler, and Alistarh]{Frantar2022GPTQAP}
Frantar, E., Ashkboos, S., Hoefler, T., and Alistarh, D.
\newblock Gptq: Accurate post-training quantization for generative pre-trained transformers.
\newblock \emph{ArXiv}, abs/2210.17323, 2022.
\newblock URL \url{https://api.semanticscholar.org/CorpusID:253237200}.

\bibitem[Han et~al.(2015)Han, Mao, and Dally]{han2015deep}
Han, S., Mao, H., and Dally, W.~J.
\newblock Deep compression: Compressing deep neural networks with pruning, trained quantization and huffman coding.
\newblock \emph{arXiv preprint arXiv:1510.00149}, 2015.

\bibitem[Hu et~al.(2021)Hu, Shen, Wallis, Allen-Zhu, Li, Wang, and Chen]{Hu2021LoRALA}
Hu, J.~E., Shen, Y., Wallis, P., Allen-Zhu, Z., Li, Y., Wang, S., and Chen, W.
\newblock Lora: Low-rank adaptation of large language models.
\newblock \emph{ArXiv}, abs/2106.09685, 2021.
\newblock URL \url{https://api.semanticscholar.org/CorpusID:235458009}.

\bibitem[Hubara et~al.(2021{\natexlab{a}})Hubara, Chmiel, Island, Banner, Naor, and Soudry]{Hubara2021AcceleratedSN}
Hubara, I., Chmiel, B., Island, M., Banner, R., Naor, S., and Soudry, D.
\newblock Accelerated sparse neural training: A provable and efficient method to find n: M transposable masks.
\newblock \emph{ArXiv}, abs/2102.08124, 2021{\natexlab{a}}.
\newblock URL \url{https://api.semanticscholar.org/CorpusID:231934142}.

\bibitem[Hubara et~al.(2021{\natexlab{b}})Hubara, Nahshan, Hanani, Banner, and Soudry]{Hubara2021AccuratePT}
Hubara, I., Nahshan, Y., Hanani, Y., Banner, R., and Soudry, D.
\newblock Accurate post training quantization with small calibration sets.
\newblock In \emph{International Conference on Machine Learning}, 2021{\natexlab{b}}.
\newblock URL \url{https://api.semanticscholar.org/CorpusID:235825979}.

\bibitem[Jaiswal et~al.(2023)Jaiswal, Gan, Du, Zhang, Wang, and Yang]{jaiswal2023compressing}
Jaiswal, A., Gan, Z., Du, X., Zhang, B., Wang, Z., and Yang, Y.
\newblock Compressing llms: The truth is rarely pure and never simple, 2023.

\bibitem[Joshi et~al.(2017)Joshi, Choi, Weld, and Zettlemoyer]{joshi2017triviaqa}
Joshi, M., Choi, E., Weld, D.~S., and Zettlemoyer, L.
\newblock Triviaqa: A large scale distantly supervised challenge dataset for reading comprehension, 2017.

\bibitem[Lester et~al.(2021)Lester, Al-Rfou, and Constant]{Lester2021ThePO}
Lester, B., Al-Rfou, R., and Constant, N.
\newblock The power of scale for parameter-efficient prompt tuning.
\newblock In \emph{Conference on Empirical Methods in Natural Language Processing}, 2021.
\newblock URL \url{https://api.semanticscholar.org/CorpusID:233296808}.

\bibitem[Li \& Liang(2021)Li and Liang]{Li2021PrefixTuningOC}
Li, X.~L. and Liang, P.
\newblock Prefix-tuning: Optimizing continuous prompts for generation.
\newblock \emph{Proceedings of the 59th Annual Meeting of the Association for Computational Linguistics and the 11th International Joint Conference on Natural Language Processing (Volume 1: Long Papers)}, abs/2101.00190, 2021.
\newblock URL \url{https://api.semanticscholar.org/CorpusID:230433941}.

\bibitem[Merity et~al.(2016)Merity, Xiong, Bradbury, and Socher]{merity2016pointer}
Merity, S., Xiong, C., Bradbury, J., and Socher, R.
\newblock Pointer sentinel mixture models, 2016.

\bibitem[OpenAI(2023)]{openai2023gpt4}
OpenAI.
\newblock Gpt-4 technical report, 2023.

\bibitem[Paperno et~al.(2016)Paperno, Kruszewski, Lazaridou, Pham, Bernardi, Pezzelle, Baroni, Boleda, and Fernández]{paperno_denis_2016_2630551}
Paperno, D., Kruszewski, G., Lazaridou, A., Pham, Q.~N., Bernardi, R., Pezzelle, S., Baroni, M., Boleda, G., and Fernández, R.
\newblock The lambada dataset, August 2016.
\newblock URL \url{https://doi.org/10.5281/zenodo.2630551}.

\bibitem[PENG et~al.(2023)PENG, Xing, Choubey, Wu, and Xiong]{peng2023model}
PENG, X., Xing, C., Choubey, P.~K., Wu, C.-S., and Xiong, C.
\newblock Model ensemble instead of prompt fusion: a sample-specific knowledge transfer method for few-shot prompt tuning.
\newblock In \emph{The Eleventh International Conference on Learning Representations}, 2023.
\newblock URL \url{https://openreview.net/forum?id=p0yrSRbN5Bu}.

\bibitem[Raffel et~al.(2020)Raffel, Shazeer, Roberts, Lee, Narang, Matena, Zhou, Li, and Liu]{2020t5}
Raffel, C., Shazeer, N., Roberts, A., Lee, K., Narang, S., Matena, M., Zhou, Y., Li, W., and Liu, P.~J.
\newblock Exploring the limits of transfer learning with a unified text-to-text transformer.
\newblock \emph{Journal of Machine Learning Research}, 21\penalty0 (140):\penalty0 1--67, 2020.
\newblock URL \url{http://jmlr.org/papers/v21/20-074.html}.

\bibitem[Sakaguchi et~al.(2019)Sakaguchi, Bras, Bhagavatula, and Choi]{sakaguchi2019winogrande}
Sakaguchi, K., Bras, R.~L., Bhagavatula, C., and Choi, Y.
\newblock Winogrande: An adversarial winograd schema challenge at scale, 2019.

\bibitem[Touvron et~al.(2023)Touvron, Lavril, Izacard, Martinet, Lachaux, Lacroix, Rozi{\`e}re, Goyal, Hambro, Azhar, Rodriguez, Joulin, Grave, and Lample]{Touvron2023LLaMAOA}
Touvron, H., Lavril, T., Izacard, G., Martinet, X., Lachaux, M.-A., Lacroix, T., Rozi{\`e}re, B., Goyal, N., Hambro, E., Azhar, F., Rodriguez, A., Joulin, A., Grave, E., and Lample, G.
\newblock Llama: Open and efficient foundation language models.
\newblock \emph{ArXiv}, abs/2302.13971, 2023.
\newblock URL \url{https://api.semanticscholar.org/CorpusID:257219404}.

\bibitem[Welbl et~al.(2017)Welbl, Liu, and Gardner]{Welbl2017CrowdsourcingMC}
Welbl, J., Liu, N.~F., and Gardner, M.
\newblock Crowdsourcing multiple choice science questions.
\newblock \emph{ArXiv}, abs/1707.06209, 2017.
\newblock URL \url{https://api.semanticscholar.org/CorpusID:1553193}.

\bibitem[Xiao et~al.(2022)Xiao, Lin, Seznec, Demouth, and Han]{Xiao2022SmoothQuantAA}
Xiao, G., Lin, J., Seznec, M., Demouth, J., and Han, S.
\newblock Smoothquant: Accurate and efficient post-training quantization for large language models.
\newblock \emph{ArXiv}, abs/2211.10438, 2022.
\newblock URL \url{https://api.semanticscholar.org/CorpusID:253708271}.

\bibitem[Xu et~al.(2023)Xu, Liu, Chen, Tang, Wang, Zhou, Hu, and Shrivastava]{Xu2023CompressTP}
Xu, Z., Liu, Z., Chen, B., Tang, Y., Wang, J., Zhou, K., Hu, X., and Shrivastava, A.
\newblock Compress, then prompt: Improving accuracy-efficiency trade-off of llm inference with transferable prompt.
\newblock \emph{ArXiv}, abs/2305.11186, 2023.
\newblock URL \url{https://api.semanticscholar.org/CorpusID:258823240}.

\bibitem[Yao et~al.(2022)Yao, Aminabadi, Zhang, Wu, Li, and He]{yao2022zeroquant}
Yao, Z., Aminabadi, R.~Y., Zhang, M., Wu, X., Li, C., and He, Y.
\newblock Zeroquant: Efficient and affordable post-training quantization for large-scale transformers, 2022.

\bibitem[Zellers et~al.(2019)Zellers, Holtzman, Bisk, Farhadi, and Choi]{zellers2019hellaswag}
Zellers, R., Holtzman, A., Bisk, Y., Farhadi, A., and Choi, Y.
\newblock Hellaswag: Can a machine really finish your sentence?, 2019.

\bibitem[Zhang et~al.(2022)Zhang, Roller, Goyal, Artetxe, Chen, Chen, Dewan, Diab, Li, Lin, Mihaylov, Ott, Shleifer, Shuster, Simig, Koura, Sridhar, Wang, and Zettlemoyer]{Zhang2022OPTOP}
Zhang, S., Roller, S., Goyal, N., Artetxe, M., Chen, M., Chen, S., Dewan, C., Diab, M.~T., Li, X., Lin, X.~V., Mihaylov, T., Ott, M., Shleifer, S., Shuster, K., Simig, D., Koura, P.~S., Sridhar, A., Wang, T., and Zettlemoyer, L.
\newblock Opt: Open pre-trained transformer language models.
\newblock \emph{ArXiv}, abs/2205.01068, 2022.
\newblock URL \url{https://api.semanticscholar.org/CorpusID:248496292}.

\end{thebibliography}
